\documentclass{article}

\usepackage{arxiv}
\usepackage{wrapfig}

\usepackage[utf8]{inputenc} 
\usepackage[T1]{fontenc}    
\usepackage{hyperref}       
\usepackage{url}            
\usepackage{booktabs}       
\usepackage{nicefrac}       
\usepackage{microtype}      
\usepackage{lipsum}
\usepackage{graphicx}
\graphicspath{ {./images/} }
\usepackage{amsmath,amsfonts,amssymb,amsthm}

\begin{document}

\title{SpikePropamine: Differentiable Plasticity in Spiking Neural Networks} 

\author{
 Samuel Schmidgall, Julia Ashkanazy, Wallace Lawson, Joe Hays\\
  U.S. Naval Research Laboratory \\
}
\date{}
\maketitle

\begin{abstract}

The adaptive changes in synaptic efficacy that occur between spiking neurons have been demonstrated to play a critical role in learning for biological neural networks. Despite this source of inspiration, many learning focused applications using Spiking Neural Networks (SNNs) retain static synaptic connections, preventing additional learning after the initial training period. Here, we introduce a framework for simultaneously learning the underlying fixed-weights and the rules governing the dynamics of synaptic plasticity and neuromodulated synaptic plasticity in SNNs through gradient descent. We further demonstrate the capabilities of this framework on a series of challenging benchmarks, learning the parameters of several plasticity rules including BCM, Oja's, and their respective set of neuromodulatory variants. The experimental results display that SNNs augmented with differentiable plasticity are sufficient for solving a set of challenging temporal learning tasks that a traditional SNN fails to solve, even in the presence of significant noise. These networks are also shown to be capable of producing locomotion on a high-dimensional robotic learning task, where near-minimal degradation in performance is observed in the presence of novel conditions not seen during the initial training period.

\end{abstract}

\section{Introduction \& Related Work}

The dynamic modification of neuronal properties underlies the basis of learning, memory, and adaptive behavior in biological neural networks. The changes in synaptic efficacy that occur on the connections between neurons play an especially vital role. 
This process, termed synaptic plasticity, is largely mediated by the interaction of pre- and post-synaptic activity between two synaptically connected neurons in conjunction with local and global modulatory signals. Importantly, synaptic plasticity is largely believed to be one of the primary bases for enabling both stable long-term learning and adaptive short-term responsiveness to novel stimuli \cite{plasticity_and_memory, Liu2012, annurev.physiol.64.092501.114547}.

An additional mechanism that guides these changes is neuromodulation. Neuromodulation, as the name suggests, is the process by which select neurons modulate the activity of other neurons; this is accomplished by the use of chemical messaging signals. Such messages are mediated by the release of chemicals from neurons themselves, often using one or more stereotyped signals to regulate diverse populations of neurons. Dopamine, a neuromodulator commonly attributed to learning \cite{Frank1940, Schultz1593, dopamine_hosp}, has been experimentally shown to portray a striking resemblance to the Temporal-Difference (TD) reward prediction error \cite{Montague1996, Schultz1593, dopamine_td} and more recently distributional coding methods of reward prediction \cite{Dabney2020DistributionalCode}. Such signals have been shown to play a critical role in guiding the effective changes in synaptic plasticity, allowing the brain to regulate the location and scale with which such changes are made \cite{Gerstner2018}. The conceptual role of dopamine has largely shaped the development of modern reinforcement learning (RL) algorithms, enabling the impressive accomplishments seen in recent literature \cite{MnihKSGAWR13, BellemareDM17, abs-1801-01290SOFTACTORCRITIC, abs-1804-08617D4PG}. While dopamine has primarily taken the spotlight in RL, many other important neuromodulatory signals have largely been excluded from learning algorithms in artificial intelligence (AI). For example, acetylcholine has been shown to play a vital role in motor control, with neuromodulatory signals often sent as far as from the brainstem to motor neurons \cite{Zaninetti1999PresenceOF}. Modeling how these neuromodulatory processes develop, as well as how neurons can directly control neuromodulatory signals are likely critical steps toward successfully reproducing the impressive behaviors exhibited by the brain. 
\nocite{adaptiverl}

Both historically and recently, neuroscience and AI have had a fruitful relationship, with neuroscientific speculations being validated through AI, and advancements in the capabilities of AI being a result of a better understanding of the brain.

A major contributor toward enabling this, particularly in AI, is through the application of backpropagation for learning the weights of Artificial Neural Networks (ANNs). Although backpropagation is largely believed to be biologically implausible \cite{bengio2015towards}, networks trained under certain conditions using this algorithm have been shown to display behavior remarkably similar to biological neural networks \cite{Banino2018, cueva2018emergence}. 

The promising advances toward more brain-like computations have led to the development of SNNs. These networks more closely resemble the dynamics of biological neural networks by storing and integrating membrane potential to produce binary spikes. Consequently, such networks are naturally suited toward solving temporally extended tasks, as well as producing many of the desirable benefits seen in biological networks such as energy efficiency, noise robustness, and rapid inference \cite{chal}. However, until recently, the successes of SNNs have been overshadowed by the accomplishments of ANNs. This is primarily due to the use of spikes for information transmission, which does not naturally lend itself toward being used with backpropagation. To circumvent this challenge, a wide variety of learning algorithms have been proposed including Spike-Timing Dependent Plasticity (STDP) \cite{abs-1804-00227, neco200806-08-804, KHERADPISHEH201856, NECO2}, ANN to SNN conversion methods \cite{fnins.2017.00682, dresabs-1805-01352, 7280696}, Eligibility Traces \cite{EPROPmaass}, and Evolutionary Strategies \cite{1556240, fnins.2014.00010, 7727228}. However, a separate body of literature enables the use of backpropagation directly with SNNs typically through the use of surrogate gradients \cite{BOHTE200217, 1abs-1202-2249, fnins.2016.00508, Slayer}. These surrogate gradient methods are primary contributors for many of the state-of-the-art results obtained using SNNs from supervised learning to RL. Counter to biology, temporal learning tasks such as RL interact with an external environment over multiple episodes before synaptic weight updates are computed. Between these update intervals, the synaptic weights remain unchanged, diminishing the potential for online learning to occur. Recent work by \cite{miconiDiffPlast} transcends this dominant fixed-weight approach specifically for the recurrent weights of ANNs by presenting a framework for augmenting traditional fixed-weight networks with Hebbian plasticity, where backpropagation updates both the weights and parameters guiding plasticity. In follow-up work, this hybrid framework was expanded to include neuromodulatory signals, whose parameters were also learned using backpropagation \cite{miconi2018backpropamine}. 

Learning-to-learn, or meta-learning, is the capability to learn or improve one's own learning ability. The brain is constantly modifying and improving its own ability to learn at both the local and global scale. This was originally theorized to be a product of neurotransmitter distribution from the Basal Ganglia \cite{DOYA2002495}, but has also included contributions from the Prefrontal Cortex \cite{Wang2018} and the Cerebellum \cite{cerebell} to name a few. In machine learning, meta-learning approaches aim to improve the learning algorithm itself rather than retaining a static learning process \cite{hospedales2020meta}. For spiking neuro-controllers, learning-to-learn through the discovery of synaptic plasticity rules offline provides a mechanism for learning on-chip since neuromorphic hardware is otherwise incompatible with on-chip backpropagation. Many neuromorphic chips provide a natural mechanism for incorporating synaptic plasticity \cite{davies2018loihi,van2018performance}, and more recently, neuromodulatory signals \cite{mikaitis2018neuromodulated}.

Despite the prevalence of plasticity in biologically-inspired learning methods, a method for learning both the underlying weights and plasticity parameters using gradient descent has yet to be proposed for SNNs. Building off of \cite{miconi2018backpropamine}, which was focused on ANNs, this paper provides a framework for incorporating plasticity and neuromodulation with SNNs trained using gradient descent. In addition, five unique plasticity rules inspired by the neuroscientific literature are introduced. A series of experiments are conducted with using a complex cue-association environment, as well as a high-dimensional robotic locomotion task. From the experimental results, networks endowed with plasticity on only the forward propagating weights, with no recurrent self-connections, are shown to be sufficient for solving challenging temporal learning tasks that a traditional SNN fails to solve, even while experiencing significant noise perturbations. Additionally, these networks are much more capable of adapting to conditions not seen during training, and in some cases displaying near-minimal degradation in performance.

\section{Differentiable Plasticity}
Section 2.1 begins by describing the dynamics of an SNN. Using these dynamic equations, Section 2.2 then introduces the generalized framework for differentiable plasticity of an SNN as well as some explicit forms of differentiable plasticity rules. First, the Differentiable Plasticity (DP) form of Linear Decay is introduced, primarily due to the conceptual simplicity of its formulation. Next, the DP form of Oja's rule \cite{oja1983subspace} is presented as DP-Oja's. This rule is introduced primarily because, unlike Linear Decay, it provides a natural and simple mechanism for stable learning, namely a penalty on weight-growth. The next form is based on the Bienenstock, Cooper, and Munro (BCM) rule \cite{Bienenstock32}, named DP-BCM. Like Oja's rule, the BCM rule provides stability, except in this case the penalty accounts for a given neuron's deviation from the average spike-firing rate. Finally, a respective set of neuromodulatory variants for the Oja's and BCM differentiable plasticity rules are presented in Section 2.3, as well as a generalized framework for differentiable neuromodulation. The rules described in this section serve primarily to demonstrate an explicit implementation of the generalized framework on two well-studied synaptic learning rules.

\subsection{Spiking Neural Network}
We will begin by describing the dynamics of an SNN, and then proceed in the following sections to describe a set of plasticity rules that can be applied to such networks. We begin with the following set of equations:

\begin{equation} \label{eq:SNN1}
\textbf{\textit{a}}^{(l)}(t) = \varepsilon*\textbf{\textit{s}}^{(l-1)}(t)
\end{equation}
\begin{equation}\label{eq:SNN2}
\textbf{\textit{u}}^{(l)}(t) = \textbf{\textit{W}}^{(l)}\textbf{\textit{a}}^{(l)}(t) + \textit{v}*\textbf{\textit{s}}^{(l)}(t)
\end{equation}
\begin{equation}\label{eq:SNN3}
\textbf{\textit{s}}^{(l)}(t) = f_{s}(\textbf{\textit{u}}^{(l)}(t)).
\vspace{1.3mm}
\end{equation}

The superscript $l \in \mathbb{N}$ represents the index for a layer of neurons and $t \in \mathbb{N}$ discrete time. We further define $n^{(l)} \in \mathbb{N}$ to represent the number of neurons in layer $l$. From here, $\varepsilon(\cdot)$ is a spike response kernel which is used to generate a spike response signal, $\textbf{\textit{a}}^{(l)}(t) \in \mathbb{R}^{n^{(l-1)}}$, by convolving incoming spikes $\textbf{\textit{s}}^{(l-1)}(t) \in \mathbb{B}^{n^{(l-1)}}, \mathbb{B} = \{0, 1\}$ over $\varepsilon(\cdot)$. We further define the binary vector $\textbf{\textit{s}}^{(0)}(t)$ to represent sensory input obtained from the environment. Often in practice, the effect of $\varepsilon(\cdot)$ provides an exponentially decaying contribution over time, which consequently has minimal influence after a fixed number of steps. Exploiting this concept, $\varepsilon(\cdot)$ may be represented as a finite weighted decay kernel with dimensionality $K$, which is chosen heuristically as the point in time with which $\varepsilon(\cdot)$ has minimal practical contribution. $\textit{v}(\cdot)$ is chosen in a similar manner, where $\textit{v}(\cdot)$ is the refractory kernel, convolving together with spikes $\textbf{\textit{s}}^{(l)}(t)$ to produce the refractory response $\textit{v}*\textit{\textbf{s}}^{(l)}(t) \in \mathbb{R}^{n^{(l)}}$. Additionally, $\textbf{W}^{(l)} \in \mathbb{R}^{n^{(l)} \times n^{(l-1)}}$ numerically represents the synaptic strength between each neuron connected from layer $l-1$ and $l$, which, as a weight is multiplied by $\textbf{\textit{a}}^{(l)}(t)$ and further summed with the refractory response $\textit{v}*\textbf{\textit{s}}^{(l)}(t)$ to produce the membrane potential $\textbf{\textit{u}}^{(l)}(t) \in \mathbb{R}^{n^{(l)}}$. The membrane potential stores the weighted sum of spiking activity arriving from incoming synaptic connections, referred to as the \textit{Post Synaptic Potential (PSP)}.

\vspace{1.3mm}

We further define the spike function $f_{s}(\cdot)$ as:

\begin{equation} \label{eq:SNN4}
f_{s}(\textit{u}): u \rightarrow s := s(t) + \delta(t-t^{(f)})
\end{equation}
\begin{equation} \label{eq:SNN5}
t^{(f)} = \min\{t: u(t) = \vartheta, t > t^{(f-1)}\}
\vspace{1.3mm}
\end{equation}

In these equations, the function $f_{s}(\cdot)$ produces a binary spike based on the neuron's internal membrane potential, $\textbf{\textit{u}}_{i}(t)$, $i \in \mathbb{N}$ indexing an individual neuron. When $\textbf{\textit{u}}_{i}(t)$ passes a threshold $\vartheta \in \mathbb{R}$, the respective binary spike is propagated downstream to a set of connected neurons, and the internal membrane potential for that neuron is reset to a baseline value $u_{r} \in \mathbb{R}$, which is often set to zero. The function enabling this is referred to as a dirac-delta, $\delta(t)$, which produces a binary output of one when $t=0$ and zero otherwise. Here, $t^{f} \in \mathbb{R}$ denotes the firing time of the $f^{th}$ spike, so that when $t = t^{(f)}$ then $\delta(t-t^{(f)}) = 1$.

\vspace{1.3mm}
Like an artificial neural network, $f_{s}(\cdot)$ can be viewed as having similar functionality to an arbitrary non-linear activation function $\phi(\cdot)$. Unlike the ANN however, $f_{s}(\cdot)$ has an undefined derivative making the gradient computation for backpropagation particularly challenging. To enable backpropagation through the non-differentiable aspects of the network, the Spike Layer Error Reassignment in Time (SLAYER) algorithm is used \cite{Slayer}. SLAYER overcomes such difficulties by representing the derivative of a spike as a surrogate gradient and uses a temporal credit assignment policy for backpropagating error to previous layers. Although SLAYER was used in this paper, we note that any spike-derivative approximation method will work together with our methods.

\subsection{Spike-based Differentiable Plasticity}

To enable differentiable plasticity we utilize the SNN dynamic equations described in (\ref{eq:SNN1}-\ref{eq:SNN5}), however now both the weights and the rules governing plasticity are optimized by gradient descent. This is enabled through the addition of a synaptic trace variable, $\textbf{\textit{E}}^{(l)}(t) \in \mathbb{R}^{n^{(l)} \times n^{(l-1)}}$, which accumulates traces of the local synaptic activities between pre- and post-synaptic connections. An additional plasticity coefficient, $\boldsymbol\alpha^{(l)} \in \mathbb{R}^{n^{(l)} \times n^{(l-1)}}$, is often learned which serves to element-wise scale the magnitude and direction of the synaptic traces independently from the trace dynamics. By augmenting our SNN we obtain:

\begin{equation}\label{eq:DPSNN1}
\textit{\textbf{a}}^{(l)}(t) = (\varepsilon*\textit{\textbf{s}}^{(l-1)}(t))
\end{equation}
\begin{equation}\label{eq:DPSNN2}
\textit{\textbf{u}}^{(l)}(t) = (\textit{\textbf{W}}^{(l)} + \boldsymbol\alpha^{(l)}\odot \textbf{\textit{E}}^{(l)}(t))\textit{\textbf{a}}^{(l)}(t) + (\textit{v}*\textit{\textbf{s}}^{(l)})(t)
\end{equation}
\begin{equation}\label{eq:DPSNN3}
\textit{\textbf{s}}^{(l)}(t) = f_{s}(\textit{\textbf{u}}^{(l)}(t)).
\vspace{1.3mm}
\end{equation}

The Hadamard product, $\odot$, is used to represent element-wise multiplication. The primary modifications from the fixed-weight SNN framework in (\ref{eq:SNN1}-\ref{eq:SNN3}) are in the addition of the synaptic trace $\textit{\textbf{E}}^{(l)}(t)$ and plasticity coefficient $\boldsymbol\alpha^{(l)}$ in (\ref{eq:DPSNN2}). Without this modification, the underlying weight $\textit{\textbf{W}}^{(l)}$ remains constant from episode-to-episode in the same way as (\ref{eq:SNN2}). However, the additional contribution of the synaptic trace $\textbf{\textit{E}}^{(l)}(t)$ enables each weight value to be modified through the interaction of \textit{local} or \textit{global} activity. The differentiable plasticity and neuromodulated plasticity frameworks presented in this work are concerned with learning such local and global signals respectively. Proceeding, we present three different methods of synaptic plasticity followed by a section describing neuromodulated plasticity. We additionally note that this framework is not limited to these particular plasticity rules and can be expanded upon to account for a wide variety of different methods.

\noindent
\textbf{2.2.1\space\space Generalized}

Neuronal activity can be represented by a diverse family of forms. Plasticity rules have been proposed using varying levels of abstraction, from spike rates and spike timing, all the way to modelling calcium-dependent interactions. To encapsulate this wide variety in our work, we abstractly define a vector $\boldsymbol\rho^{(l)}(t)$ to represent activity for a layer of neurons $l$ at time $t$. In many practical instances, time $t$ may represent continuous time, however our examples and discussions are primarily concerned with the evolution of discrete time, which is the default mode from which many models of spiking neurons operate. Likewise, the activity vector $\boldsymbol\rho^{(l)}(t)$ may be represented by a variety of different sets, such as $\mathbb{B}^{n^{(l)}}$ or $\mathbb{R}^{n^{(l)}}$ for spike-timing or rate-based activity; this is primarily dependent on which types of activity the experimenter desires to model. The generalized equation for differentiable plasticity is expressed as follows:

\begin{equation}\label{eq:DPSNNGEN}
\textbf{\textit{E}}^{(l)}(t+\Delta\tau) = F(\boldsymbol\rho^{(l-1)}(t), \boldsymbol\rho^{(l)}(t), \textbf{\textit{E}}^{(l)}(t), \textit{L}^{(l)}).
\end{equation}

Here, $\textbf{\textit{E}}^{(l)}(t+\Delta\tau)$ is updated after a specified time interval $\Delta\tau \in \mathbb{N}$. In these equations, $F(\cdot)$ is a function of the pre- and post-synaptic activity, $\boldsymbol\rho^{(l-1)}(t)$ and $\boldsymbol\rho^{(l)}(t)$, as well as $\textbf{\textit{E}}^{(l)}(t)$ at the current time-step and $\textit{L}^{(l)}$ which represents an arbitrary set of functions describing \textit{local} neuronal activity from either pre- or post-synaptic neurons. In practice, $\textbf{\textit{E}}^{(l)}(t=0)$ is often set to zero at the beginning of a new temporal interaction.

\noindent
\textbf{2.2.2\space\space DP-Linear Decay}

Perhaps the simplest form of differentiable plasticity is the linear decay method:

\begin{equation}\label{eq:DPSNNLinear}
\textbf{\textit{E}}^{(l)}(t+\Delta\tau) = (1-\eta^{(l)})\textbf{\textit{E}}^{(l)}(t) + \eta^{(l)}(\boldsymbol\rho^{(l)}(t))^\intercal\boldsymbol\rho^{(l-1)}(t).
\end{equation}

Let the set $\textit{L}^{(l)} = \{\eta^{(l)}\}$, with $\eta^{(l)} \in \mathbb{R}$. In this equation, $\textbf{\textit{E}}^{(l)}(t+\Delta\tau)$ is computed using the local layer-specific function $\eta^{(l)}$, representing the rate at which new local activity $\boldsymbol\rho^{(l)}(t)(\boldsymbol\rho^{(l-1)}(t))^\intercal \in \mathbb{R}^{n^{(l)} \times n^{(l-1)}}$ is incorporated into the synaptic trace, as well as the degree to which prior synaptic activity will be 'remembered' from $(1-\eta^{(l)})\textbf{\textit{E}}^{(l)}(t)$. While the parameters regulating $\textbf{\textit{E}}^{(l)}(t+\Delta\tau)$ will generally approach values that produce stable weight growth, in practice $\textbf{\textit{E}}^{(l)}(t)$ is often clipped to enforce stable bounds. Here, the local variable $\eta^{(l)}$ acts as a free parameter and is learned through gradient descent.

\noindent
\textbf{2.2.3\space\space DP-Oja's}

Among the most studied synaptic learning rules, Oja's rule simplistically provides a natural system of stability and effective correlation \cite{Oja1982}. This rule balances potentiation and depression directly from the synaptic activity stored in the trace, which cause a decay proportional to its magnitude. Mathematically, Oja's rule enables the neuron to perform Principal Component Analysis (PCA) which is a common method for finding unsupervised statistical trends in data \cite{oja1983subspace}. Building off of this work, we incorporate Oja's rule into our framework as Differentiable Plasticity Oja's rule (DP-Oja's). Rather than a generalized representation of activity, DP-Oja's rule uses a more specific rate-based representation, where $\boldsymbol\rho^{(l)}(t) = \textbf{\textit{r}}^{(l)}(t) \in \mathbb{R}^{n^{(l)}}$. To obtain $\textbf{\textit{r}}^{(l)}(t)$, spike averages are computed over the pre-defined interval $\Delta\tau$.
DP-Oja's rule is defined as:

\begin{equation}\label{eq:DPSNNOJA}
\textbf{\textit{E}}^{(l)}(t+\Delta\tau) = (1-\eta^{(l)})\textbf{\textit{E}}^{(l)}(t) + \eta^{(l)}(\textbf{\textit{r}}^{(l)} - \textbf{\textit{E}}^{(l)}\textbf{\textit{r}}^{(l-1)})(t)^{\intercal}\textbf{\textit{r}}^{(l-1)}(t).
\vspace{1.3mm}
\end{equation}

Similar to (\ref{eq:DPSNNLinear}), we let the set $\textit{L}^{(l)} = \{\eta^{(l)}\}$ contain the local layer-specific value $\eta^{(l)}$ that governs the incorporation of novel synaptic activity. Differing however, $\textbf{\textit{E}}^{(l)}(t)$ is used twice. The $\textbf{\textit{E}}^{(l)}(t)$ on the left-hand side serves a similar purpose compared with (\ref{eq:DPSNNLinear}), however on the right-hand side this value penalizes unbounded growth, acting as an unsupervised regulatory mechanism. Here, like in (\ref{eq:DPSNNLinear}), $\eta^{(l)}$ acts as a free parameter learned through gradient descent.

\textbf{2.2.4\space\space DP-BCM}

Another well-studied example of plasticity is the BCM rule \cite{Bienenstock32}. The BCM rule has been shown to exhibit similar behavior to STDP under certain conditions \cite{articlestdpbcm}, as well as to successfully describe the development of receptive fields \cite{Law1994, articlebcm}. BCM differs from Oja's rule in that it has more direct control over potentiation and depression through the use of a dynamic threshold which often represents the average spike rate of each neuron. In this example of differentiable plasticity, we describe a model of BCM, where the dynamics governing the plasticity as well as the stability-providing sliding threshold are learned through backpropagation, which we refer to as Differentiable Plasticity BCM (DP-BCM). This rule can be described as follows:

\begin{equation}\label{eq:DPSNNBCM}
\textbf{\textit{E}}^{(l)}(t+\Delta\tau) = (1-\eta^{(l)})\textbf{\textit{E}}^{(l)}(t) + \eta^{(l)}(\textbf{\textit{r}}^{(l)}(t))^{\intercal}\textbf{\textit{r}}_{\beta}^{(l)}(t)
\vspace{1.3mm}
\end{equation}
\begin{equation}\label{eq:DPSNNSUP}
\textbf{\textit{r}}_{\beta}^{(l)}(t) = \textbf{\textit{r}}^{(l-1)}(t)\odot(\textbf{\textit{r}}^{(l-1)}(t) - ( \boldsymbol\phi^{(l)}(t) + \boldsymbol\psi^{(l)}))
\vspace{1.3mm}
\end{equation}
\begin{equation}\label{eq:DPSNNBCMR}
\boldsymbol\phi^{(l)}(t+\Delta\tau) = (1-\eta^{(l)}_{\phi})\boldsymbol\phi^{(l)}(t) + \eta^{(l)}_{\phi}\omega(\textbf{\textit{r}}^{(l-1)}(t)).
\vspace{1.3mm}
\end{equation}

As in (\ref{eq:DPSNNOJA}), the DP-BCM uses a rate-based representation of synaptic activity, where $\boldsymbol\rho^{(l)}(t) = \textbf{\textit{r}}^{(l)}(t)$. Here, we let the set of local functions $\textit{L}^{(l)} = \{\boldsymbol\psi^{(l)}, \boldsymbol\phi^{(l)}, \eta^{(l)}_{\phi}, \eta^{(l)}\}$. To begin, $\boldsymbol\psi^{(l)} \in \mathbb{R}^{n^{(l-1)}}$ is a bias vector that remains static during interaction time, and the parameter $\boldsymbol\phi^{(l)}(t) \in \mathbb{R}^{n^{(l-1)}}$ is its dynamic counterpart. These parameters enable the addition of a sliding-boundary, $\boldsymbol\phi^{(l)}(t) + \boldsymbol\psi^{(l)}$, which determines whether activity results in potentiation versus depression. The dynamics of this boundary are described in (\ref{eq:DPSNNBCMR}). Otherwise, $\omega(\cdot)$ serves as an arbitrary function of the pre-synaptic activity $\textbf{\textit{r}}^{(l-1)}(t)$, and $\eta^{(l)}_{\phi} \in \mathbb{R}$ determines the rate at which new information is incorporated into the $\boldsymbol\phi^{(l)}(t)$ trace. For our experiments we let $\omega(\cdot) = I(\cdot)$, which is the identity function. This altogether has the effect of slowly incorporating the observed rate of pre-synaptic activity $\textbf{\textit{r}}^{(l-1)}(t)$ into $\boldsymbol\phi^{(l)}(t)$. Finally, $\eta^{(l)} \in \mathbb{R}$ provides the same utility as in (\ref{eq:DPSNNLinear}). Comparatively, the BCM rule is more naturally suited for regulating potentiation and depression than Oja's, which primarily regulates depression through synaptic weight decay. Among the local functions, $\boldsymbol\psi^{(l)}$, $\eta^{(l)}_{\phi}$, and $\eta^{(l)}$ are free parameters learned through gradient descent.

\subsection{Spike-based Differentiable Neuromodulation}

In addition to differentiable plasticity, a framework for differentiable neuromodulation is also presented. Neuromodulation, or neuromodulated plasticity, allows the use of both learned local signals, as well as learned \textit{global} signals. These signals modulate the effects of plasticity by scaling the magnitude and direction of synaptic modifications based on situational neuronal activity. This adds an additional layer of learned supervision, enabling the potential for learning-to-learn, or \textit{meta-learning}. Adding neuromodulation provides an analogy to the neuromodulatory signals observed in biological neural networks, which provide a rich set of biological processes to take inspiration from.

As before, we present the equation for generalized neuromodulated plasticity and further describe a series of more specific neuromodulation rules. 

\textbf{2.3.1\space\space Generalized Neuromodulated Plasticity}

The generalized equation for neuromodulated plasticity is as follows:
\begin{equation}\label{eq:NMSNNGEN}
\textbf{\textit{E}}^{(l)}(t+\Delta\tau) = G(\boldsymbol\rho^{(l-1)}(t), \boldsymbol\rho^{(l)}(t), \textbf{\textit{E}}^{(l)}(t), \textit{L}^{(l)}, \textit{M}).
\end{equation}
In this equation, $G(\cdot)$ has the same functionality as $F(\cdot)$ in (\ref{eq:DPSNNGEN}) except for the addition of neuromodulatory signals $\textit{M}$. Here, the values contained in $\textit{M}$ may be represented by a wide variety of functions, however it differs from $\textit{L}^{(l)}$ in that the elements may express global signals. Global signals may be computed at any part of the network, or by a separate network all together. Additionally, global signals may be incorporated that are learned independent of the modulatory reaction, such as dopamine-inspired TD-error from an independent value-prediction network, or a function computing predictive feedback-error signals as is observed in the cerebellum \cite{Popa2019}.

\textbf{2.3.2\space\space NDP-Oja's}

Building off of (\ref{eq:DPSNNOJA}), Oja's synaptic update rule is augmented with a neuromodulatory signal that linearly weights the neuronal activity of post-synaptic neurons. This neuromodulated variant of Oja's rule (NDP-Oja's) is described as follows:

\begin{equation}\label{eq:NMSNNOJA}
\textbf{\textit{E}}^{(l)}(t+\Delta\tau) = (1-\eta^{(l)})\textbf{\textit{E}}^{(l)}(t) + \eta^{(l)}(\textbf{\textit{M}}^{(l)}(t)\odot\textbf{\textit{r}}^{(l)} - \textbf{\textit{E}}^{(l)}\textbf{\textit{r}}^{(l-1)})(t))^{\intercal}\textbf{\textit{r}}^{(l-1)}(t).
\vspace{1.3mm}
\end{equation}
\begin{equation}\label{eq:NMSNNOJAM}
 \textbf{\textit{M}}^{(l)}(t) =  \textbf{\textit{W}}^{(l)}_{m} \textbf{\textit{r}}^{(l)}(t).
\vspace{1.3mm}
\end{equation}

Where the parameter $\textbf{\textit{W}}^{(l)}_{m} \in \mathbb{R}^{n^{(l)} \times n^{(l)}}$ weights the post-synaptic activity $\textbf{\textit{r}}^{(l)}(t)$, which modulates the right-hand trace dynamics in (\ref{eq:NMSNNOJA}). Importantly, the effect of the gradient in learning $\textbf{\textit{M}}^{(l)}(t)\in \mathbb{R}^{n^{(l)}}$ also contributes toward modifying the parameters producing the post-synaptic activity $\textbf{\textit{r}}^{(l)}(t)$ rather than simply having a passive relationship. This enables more deliberate and effective control of the neuromodulatory signal. In this equation, both $\textbf{\textit{W}}^{(l)}_{m}$ and $\eta^{(l)}$ are free parameters learned through gradient descent. Otherwise, the role of each parameter is identical to (\ref{eq:DPSNNOJA}). 

\textbf{2.3.3\space\space NDP-BCM}

In a similar manner, Equations (\ref{eq:DPSNNBCM}-\ref{eq:DPSNNBCMR}) for DP-BCM are augmented with a neuromodulatory signal that linearly weights the neuronal activity of post-synaptic neurons, which is referred to as Neuromodulated Differentiable Plasticity BCM (NDP-BCM). This rule can be described as follows:

\begin{equation}\label{eq:NMSNNBCM}
\textbf{\textit{E}}^{(l)}(t+\Delta\tau) = (1-\eta^{(l)})\textbf{\textit{E}}^{(l)}(t) + \eta^{(l)}(\textbf{\textit{M}}^{(l)}(t)\odot\textbf{\textit{r}}^{(l)}(t))^{\intercal}\textbf{\textit{r}}_{\beta}^{(l)}(t)
\vspace{1.3mm}
\end{equation}
\begin{equation}
\textbf{\textit{r}}_{\beta}^{(l)}(t) = \textbf{\textit{r}}^{(l-1)}(t)\odot(\textbf{\textit{r}}^{(l-1)}(t) - ( \boldsymbol\phi^{(l)}(t) + \boldsymbol\psi^{(l)}))
\vspace{1.3mm}
\end{equation}
\begin{equation}\label{eq:NMSNNBCMR}
\boldsymbol\phi^{(l)}(t+\Delta\tau) = (1-\eta^{(l)}_{\phi})\boldsymbol\phi^{(l)}(t) + \eta^{(l)}_{\phi}\omega(\textbf{\textit{r}}^{(l-1)}(t)).
\vspace{1.3mm}
\end{equation}
\begin{equation}\label{eq:NMSNNBCMM}
 \textbf{\textit{M}}^{(l)}(t) =  \textbf{\textit{W}}^{(l)}_{m} \textbf{\textit{r}}^{(l)}(t).
\vspace{1.3mm}
\end{equation}

As above, the learned parameter $\textbf{\textit{W}}^{(l)}_{m} \in \mathbb{R}^{n^{(l)} \times n^{(l)}}$ weights the post-synaptic activity $\textbf{\textit{r}}^{(l)}(t)$, which is then distributed to modulate the right-hand trace dynamics in (\ref{eq:NMSNNBCM}). Otherwise, each parameter follows from (\ref{eq:DPSNNBCM}-\ref{eq:DPSNNBCMR}). Similarly, $\boldsymbol\psi^{(l)}$, $\eta^{(l)}_{\phi}$, and $\eta^{(l)}$ as well as $\textbf{\textit{W}}^{(l)}_{m}$ are free parameters learned through gradient descent.

\section{Results}
The results of this work demonstrate the improvements in performance that differentiable plasticity provides over fixed-weight SNNs, as well as the unique behavioral patterns that emerge as a result of differentiable plasticity. Presented here are two distinct environments which require challenging credit assignment.

\subsection{Noisy Cue-Association: Temporal Credit-Assignment Task}

\begin{figure}
\begin{center}
  \makebox[\textwidth]{\includegraphics[width=0.85\paperwidth]{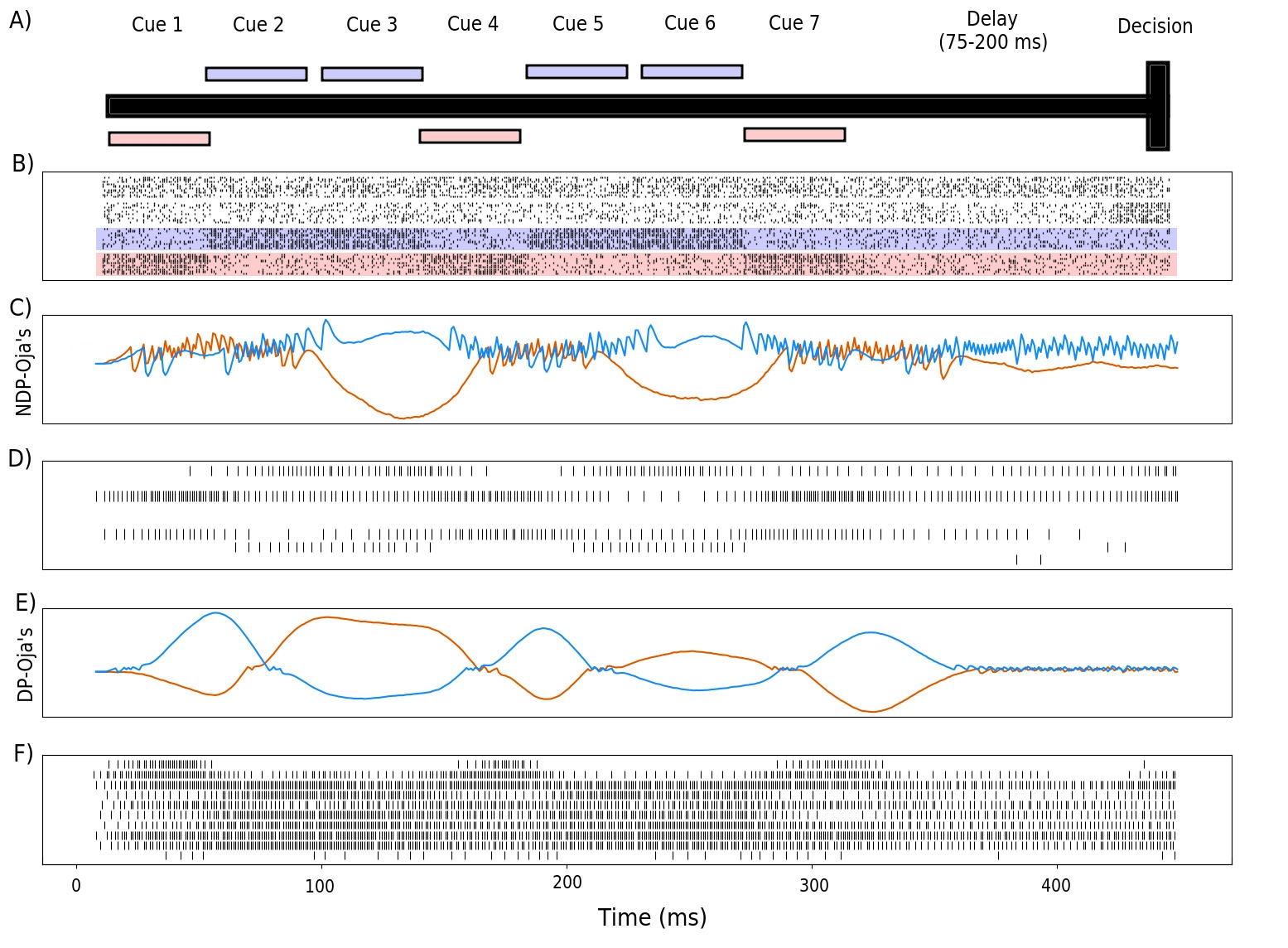}}
\end{center}
\caption{(A) A graphical visualization of the cue-association T-Maze, where the cues are sequentially presented followed by a delay and decision period. The right cue is shown in red, the left in blue. (B) Displayed is the sensory neuron input activity. From bottom to top, the activity of ten neurons each represent left and right cues, followed by ten neurons for an action decision cue, and finally ten neurons that have activity with no relationship to the task. (C-F) Membrane potential for the output neurons, and the spiking activity of a random subset of ten hidden neurons for neuromodulated Oja's rule [C, D] and non-modulated Oja's rule [E, F]. The blue and red curves here correspond to the neuron representing the decision for choosing left or right respectively, also corresponding to the left and right cue colors in [A-B]. In (D) only 5 of the 10 neurons were actively spiking, whereas in (F) all 10 were.}
\end{figure}

Experience-dependent synaptic changes provide critical functionality for both short- and long-term memory. Importantly, such a mechanism should be able to disentangle the correlations between complex sensory cues with \textit{delayed} rewards, where the learning agent often has to wait a variable amount or time before an action is made and a reward is received.
A common learning experiment in
neuroscience analyzes the performance of rodents in a similar context through the use of a T-maze training environment \cite{KUSMIERZ2017170,Engelhard2019}. Here, a rodent moves down a straight corridor where a series of sensory visual cues are arranged randomly on the left and right of the rodent as it walks toward the end of the maze. After the sensory cues are displayed, there is a delay interval between the cues and the decision period. Finally, at the T-junction, the rodent is faced with the decision of turning either left of right. A positive reward is given if the rodent chooses the side with the highest number of visual cues. This environment poses unique challenges representative of a natural temporal learning problem, as the decision-making agent is required to learn that reward is independent of both the temporal order of each cue as well as the side of the final cue.

\begin{figure}%
    \centering
    \includegraphics[width=17.5cm]{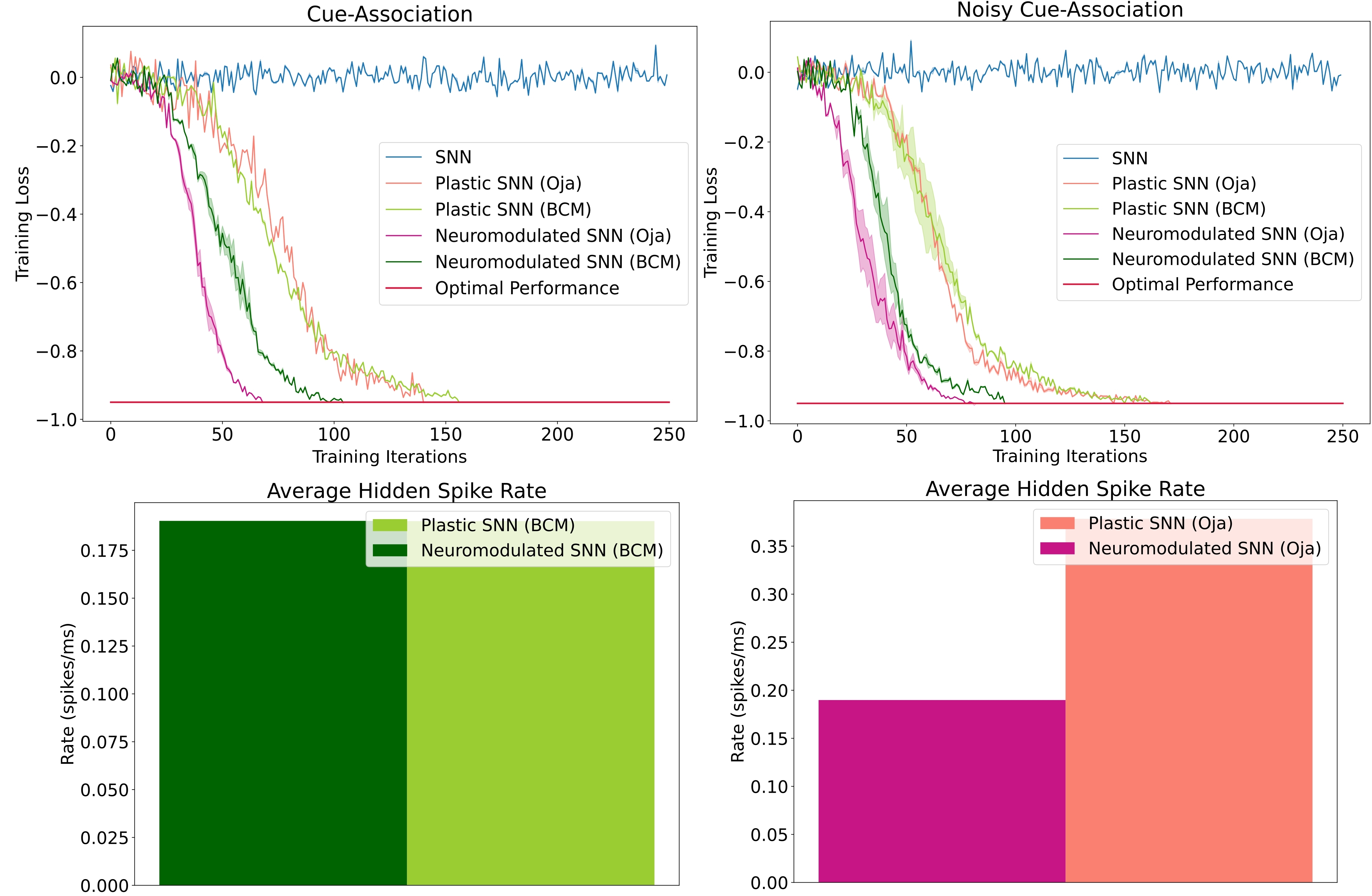}
    \caption{(A-B) The performance of Oja's and BCM plastic and neuromodulated learning rules on the low-noise (A) and high-noise (B) environments. (C-D) The average spike firing rate of hidden layer neurons in the high-noise environments for plastic and neuromodulated Oja's (C) and BCM (D). Once the average network training loss is less than -0.97, which here is consider optimal, the training is halted. Training accuracy is defined as the ratio of correct cues averaged over a series of trials. To perform gradient descent, training loss is this value multiplied by $-1$.}
\end{figure}

Rather than visual cues specifically, the cues in our experiment produce a time period of high-spiking activity that is input into a distinct set of neurons for each cue (Figure 1). Additionally, a similarly-sized set of neurons begins producing spikes near the T-junction indicating a decision period, from which the agent is expected to produce a decision to go left or right. Finally, the last set of neurons produce noise to make the task more challenging. The cue-association task has been shown to be solvable in simulation with the use of recurrent spiking neural networks for both Backpropagation Through Time (BPTT) and eligibility propagation (E-Prop) algorithms \cite{EPROPmaass}. However, using the same training methods, a feedforward spiking neural network without recurrent connections is not able to solve this task. In the \cite{miconi2018backpropamine} experiments, results were shown for ANNs with plastic and neuromodulated synapses on only the recurrent weights.  In this experiment, we determine whether non-recurrent feedforward networks with plastic synapses are sufficient for solving this same task. Here, we consider both DP-BCM 
and DP-Oja's rules as well as the respective neuromodulatory variants described in Sections 2.2-2.3. We also collect results from an additional environment where each population of neurons has a significantly higher probability of spiking randomly, as well as a reduced probability of spiking during the actual cue interval.

The input layer is comprised of 40 neurons: 10 for the right-sided cues, 10 for the left-sided cues, 10 neurons which display activity during the decision period, and 10 neurons which produce spike noise (Figure 1 (B)). The hidden layer is comprised of 64 neurons, where each neuron is synaptically connected to every neuron in the input layer. Finally, the output layer is similarly fully-connected with two output neurons. 

Output activity is collected over the decision interval averaging the number of spikes over each distinct output neuron. To decide which action is taken at the end of the decision interval, the output activity is used as the 2-dimensional log-odds input for a binomial distribution from which an action is then sampled. To compute the parameter gradients, the policy gradient algorithm is employed together with BPTT and the Adam optimization method \cite{kingma2014adam}. To compute the policy gradient, a reward of one is given for successfully solving the task, where otherwise a reward of negative one is given. A more in-depth description of the training details is reserved for the Section 5.4 of the Appendix.

\begin{wrapfigure}{l}{0.27\textwidth}
    \centering
    \includegraphics[width=0.27\textwidth]{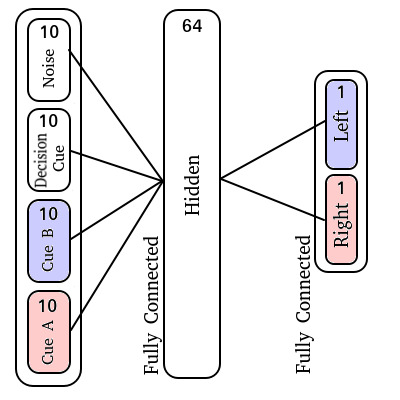}
  \caption{Cue-Association network diagram}
\end{wrapfigure}

Accuracy on this task is defined as the ratio of correct cue-decisions at the end of the maze averaged over 100 trials. Training loss is defined as accuracy multiplied by $-1$, hence the optimal performance is $-1$. The training results for both the high- and low-noise environments are shown in (Figure 2). For both environments, the NDP-BCM and NDP-Oja's consistently outperform both DP-BCM and DP-Oja's, whereas the non-plastic SNN fails to solve the task. The neuromodulated networks learn to solve the task most efficiently, despite having more complex dynamics as well as a larger set of parameters to learn. In comparing the non-modulated plasticity variants, there is minimal difference in learning efficiency for either environment. Interestingly, there was minimal degradation in training performance when transitioning from a low- to high-noise environment as shown in (Figure 2 (A-B)). 

One observed difference between the activity of the neuromodulated and non-modulated variants of BCM and Oja's rule is their average hidden spike rates. The average spike-firing rate is relatively consistent between NDP-Oja's and NDP-BCM, as well as the DP-BCM, however the DP-Oja's network has almost twice the spike-firing rate of the former three networks (Figure 2). This is thought to be due to Oja's rule primarily controlling synaptic depression through weight decay, which tends to produce larger weight values in active networks. This differs from the BCM rule, which produces a sliding boundary based on the average neuronal spike-firing rate to control potentiation and depression. The neuromodulated variant of Oja's rule however, can control potentiation and depression through the learned modulatory signal, bypassing the weight-value associated decay. This effect is also seen in (Figure 1 (D,F)) where the spiking activity of 10 randomly-sampled neurons is shown. Again, the NDP-Oja's rule shows drastically lower average spiking activity. This demonstrates that neuromodulatory signals may provide critical information in synaptic learning rules where depression is not actively controlled. Reduced activity is desirable in neuromorphic hardware, as it enables lower energy consumption in practical applications. 

To better understand the role of neuromodulation in this experiment, four unqiue cue-association cases are considered (Figure 6-9, Appendix). It is observed that the modulatory signals on the output neurons exhibit loose symmetry, whereas the activity on hidden neurons follow a similar dynamic pattern for different cues. Additionally, the plastic-weights are shown to behave differently when deprived of sensory-cues. During this deprivation period, the characteristic potentiation and depression seen in all other cue patterns is absent. It is evident from these experiments that the plasticity and neuromodulation have a significant effect on self-organization and behavior. A more in-depth analysis of the neuronal activity is provided in the Appendix (Section 5.6).

\begin{figure}%
    \centering
    \includegraphics[width=17.5cm]{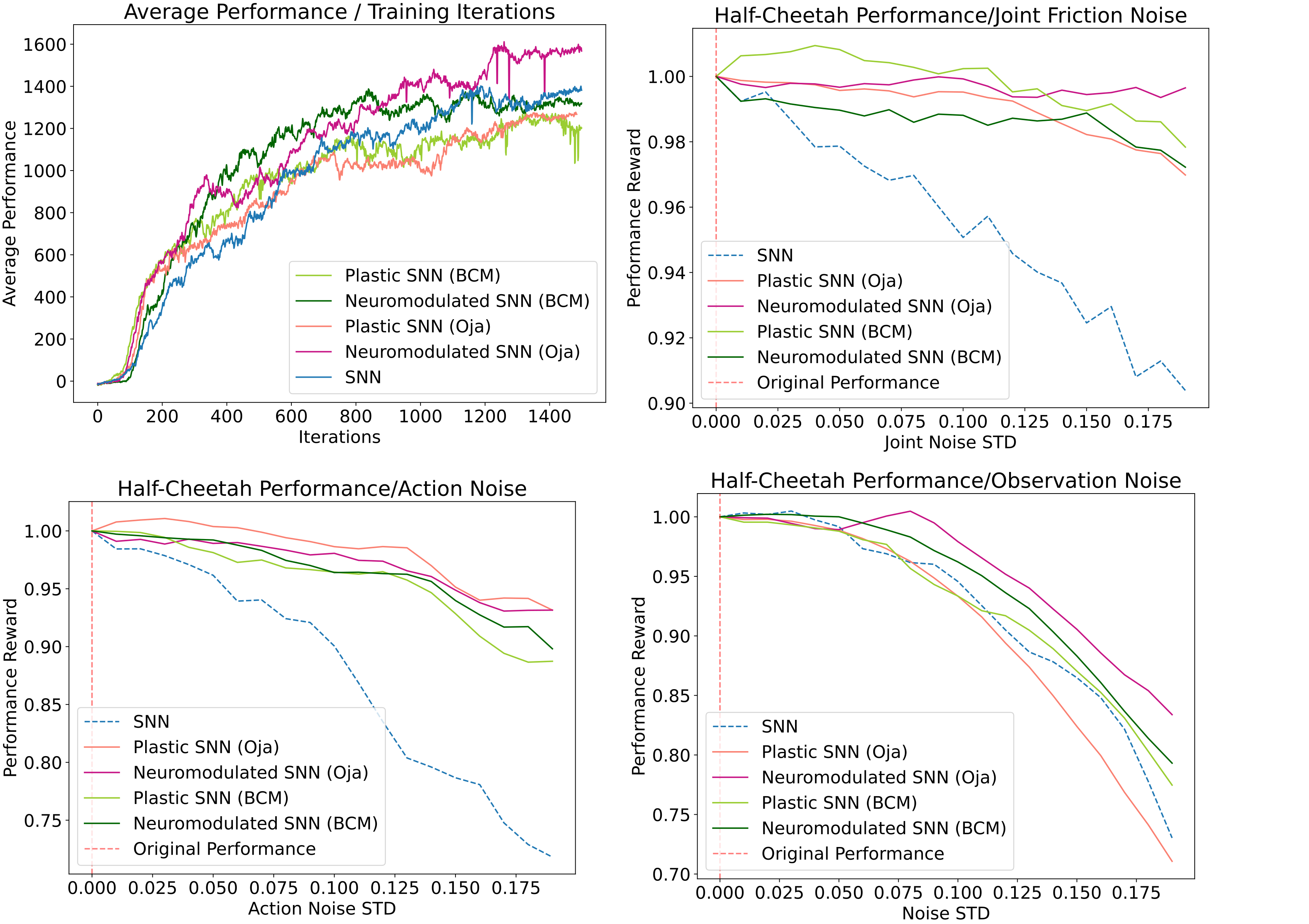}
    \caption{(E) Average performance of each network at each iteration of the training process. (F-H) Average loss in performance as a ratio of the observed performance in response to noise [x-axis] and the original network performance in the absence of noise at x=0. Averages for each noise standard deviation were collected over 100 training episodes.}
\end{figure}

\subsection{High-dimensional Robotic Locomotion Task}

Locomotion is among the most impressive capabilities of the brain. The utility of such a capability for a learning agent extends beyond biological organisms, and has received a long history of attention in the robotics field for its many practical applications. Importantly, among the most desirable properties of a locomotive robot are \textit{adaptiveness}, \textit{robustness}, as well as energy efficiency. However, it is worth noting that the importance of having adaptive capabilities for a locomotive agent primarily serves to enable robust performance in response to noise, varying environments, and novel situations. Since plasticity in networks largely serves as a mechanism toward adapting appropriately to new stimuli, we test the adaptive capabilities of our differentiable plasticity networks in a locomotive robotic learning setting.

We thus begin by considering a modified version of the Half-Cheetah environment. Half-Cheetah is a common benchmark used to examine the efficacy of RL algorithms. This environment begins with a robot which loosely has the form of a cheetah, controlling six actuated joints equally divided among the two limbs. Additionally, the robot is restricted to motion in 2-dimensions, hence the name 'Half-Cheetah'. In total, the half-cheetah is a 9-DOF system, with 3 unactuated floating body DOFs and 6 actuated-DOFs for the joints. The objective of this environment is to maximize forward velocity, while retaining energy efficiency. The sensory input for this environment is comprised of the relative angle and angular velocity of each joint for a total of 12 individual inputs. Originally, the environmental measurements are represented as floating point values. These measurements are then numerically clipped, converted into a binary spike representation, and sent as input into the network. The binary spike representation utilized is a probabilistic population representation based on place coding. Similar to the sensory representation, the action outputted by the network is represented by a population of spiking neurons. In each population there exists spiking neurons with equal sized positive and negative sub-populations. The total sum of spikes for each population is then individually collected and averaged over the pre-defined integration interval $T \in \mathbb{N}$. Both the equations describing the spike observation and action representation are further discussed in Section 5.2 and 5.3 of the Appendix.

To introduce action variance for this experiment, the output floating point value $\textbf{A}(t)$ is used as the mean for a multivariate Gaussian with zero co-variance, $\textbf{A}_{e}(t) = \mathcal{N}(\textbf{A}(t), \text{exp}(\boldsymbol\sigma_{log})^{2})$. The log standard deviation, $\boldsymbol\sigma_{log}$, is a fixed vector that is learned along with the network parameters. To produce an action, the integration interval $T$ was chosen to be 50 time-steps and the action sub-population size to be 100 neurons. With the action floating point dimensionality having been 6, this produced a spike-output dimensionality of 600 neurons. Additionally, using a population size of 50 neurons for each state input and a 12-dimensional input, the spike-input dimensionality was also 600 neurons. Each network model in this experiment is comprised of 2 fully-connected feed-forward hidden layers with 64 neurons each. 

To compute the gradients for the network parameters we used BPTT with the surrogate gradient method Proximal Policy Optimization, altogether with the Adam optimization method \cite{schulman2017proximal, kingma2014adam}. For the policy gradient, the reinforcement signal is given for each action output proportional to forward velocity, with an energy penalty on movement. This signal is backpropagated through the non-differentiable spiking neurons using SLAYER, with modifications described in Section 5.1 of the Appendix \cite{Slayer}.

\begin{wrapfigure}{l}{0.31\textwidth}
    \centering
    \includegraphics[width=0.31\textwidth]{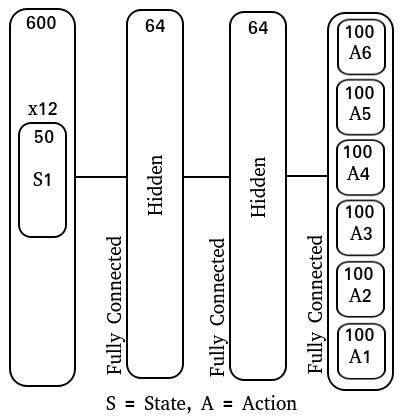}
  \caption{Locomotion network diagram}
\end{wrapfigure}

Five unique network types are evaluated on the locomotion task: a traditional SNN, the DP-BCM, DP-Oja's, NDP-BCM, and NDP-Oja's. Three unique categories of noise are measured on the trained networks: joint friction noise, action noise, and observation noise. For each of these categories, a noise vector is sampled from a Gaussian distribution with a mean of zero and a specified standard deviation. Then, to obtain an accurate measurement, the performance for an individual network is collected and the objective is averaged over 100 performance evaluations to represent the average performance response for that network at the specified noise standard deviation (Figure 4 (F-H)). Both the noise vector dimensionality and the way in which it is utilized is uniquely defined by the nature of each task. To represent observation noise, a 12-dimensional noise vector $\textbf{z}(t) \sim \mathcal{N}(\textbf{0}, \textbf{1}\sigma^{2})$, with a standard deviation $\sigma$ which remains fixed starting at the beginning of an episode, is sampled for each observation at time $t$, and further summed with the respective observation before conversion to spike representation, $\textbf{X}_{z}(t) = \textbf{X}(t) + \textbf{z}(t)$. This observation noise is in addition to the spike-firing probability noise, $\vartheta_{min}$, described in Section 5.2 of the Appendix. For the action noise, the same sampling process is repeated for a 6-dimensional vector, however here the noise vector plus one is element-wise multiplied with the output action, $\textbf{A}_{z}(t) = \textbf{A}(t) \odot (\textbf{1} + \textbf{z}(t))$. The joint friction noise rather is sampled at the beginning of a performance episode $t=0$ and held constant throughout that episode, $\textbf{z}(t) = \textbf{z}(t=0)$. The sampled noise is then further summed as a percentage of the originally specified joint friction constants $\textbf{f}_{z}(t)$ at the beginning of the performance evaluation episode, $\textbf{f}_{z}(t) = \textbf{f}(t) \odot (\textbf{1} + \textbf{z}(t))$. 

Despite each network having been trained in the absence of these noise types, the post-training performance response of the networks vary (Figure 4). Overall, the networks augmented with differentiable plasticity are shown to provide more effective adaptive capabilities, where minimal loss in performance was observed during the joint friction and action noise experiment for plastic networks (Figure 4). However, while these networks displayed improvements in robustness over the joint friction and action noise tasks, they did not display improvements on the observation noise task compared with the fixed-weight SNN. 

The activity of the modulatory signals in this task seem to noisily oscillate within a consistent range after the initial timestep (Figure 11, Appendix). This behavior is observed to be consistent across various noise perturbations, and in the absence of them. However, when deprived of sensory input, as in the case where the robot flips on its back, modulatory oscillations cease almost completely (Figure 10, Appendix). This differs from the modulatory behavior in the cue-association task where the hidden signals potentiate and decay significantly during the sensory cue sequence. A more in-depth analysis for this task is provided in the Appendix (Section 5.8). 

Training performance results for both the NDP- and DP-network variants are relatively consistent for each experiment included in (Figure 4 (E)). This differs from the cue-association experiment, where NDP-networks converged on the training task around 50 iterations faster than DP-networks. This experiment differs fundamentally in that the locomotion task is inherently solvable without temporal learning capabilities \cite{schulman2017proximal}, hence deciphering the role and benefit of plasticity and neuromodulation is not trivial. Additionally, neuromodulatory signals in biological networks do not act solely on modifying synaptic efficacy, rather have a whole host of effects depending on the signal, concentration, and region. Perhaps advances in the capabilities of NDP-networks will be a result of introducing these biologically inspired modulations \cite{Zaninetti1999PresenceOF, Dabney2020DistributionalCode, dopamine_hosp}.

\section{Discussion}

We have proposed a framework for learning the rules governing plasticity and neuromodulated plasticity, in addition to fixed network weights, through gradient descent on SNNs, providing a mechanism for online learning. Additionally, we have provided formulations for a variety of plasticity rules inspired by neuroscience literature, as well as general equations from which new plasticity rules may be defined. Using these rules, we demonstrated that synaptic plasticity is sufficient for solving a noisy and complex cue-association environment where a fixed-weight SNN fails. These networks also display an increased robustness to noise on a high-dimensional locomotion task. We also showed that the average spike-firing rate for DP-Oja's rule is reduced to the same observed rates seen in DP-BCM and NDP-BCM in the presence of a neuromodulatory signal, and hence more energy efficient. One potential limitation of this work is that, while gradients provide a strong and precise mechanism for learning in feed-forward and self-recurrent SNNs, there is no straightforward mechanism for backpropagating the gradients of feedback weights which are often incorporated in biologically-inspired network architectures. Another limitation is the computation cost associated with BPTT, which has a non-linear complexity with respect to weights and time. This limitation may be alleviated with truncated BPTT \cite{trunc}, however this reduces gradient accuracy and hence often performance as well. 

The incorporation of synaptic plasticity rules together with SNNs has a history that spans almost the same duration as SNNs themselves. The implications of a framework for learning these rules using the power of gradient descent may prove to showcase the inherent advantages that SNNs provide over ANNs on certain learning tasks. One task that may naturally benefit from this framework is in the domain of Sim2Real, where the behavior of policies learned in simulation are transferred to hardware. Often small discrepancies between a simulated environment and the real world prove too challenging for a reinforcement-trained ANN, especially on fine-motor control tasks. The improved response to noise displayed in our experimental results for DP-SNNs and NDP-SNNs on the robotic locomotion task may benefit the transfer from simulation to real hardware. Additionally, the inherent online learning capabilities of differentiable plasticity may provide a natural mechanism for on-chip learning in neurorobotic systems.

While our framework leverages the work of \cite{miconi2018backpropamine} to enter the SNN domain, this work also introduces novel results and further innovations. In the \cite{miconi2018backpropamine} experiments, results were shown for networks with plastic and neuromodulated synapses on only the recurrent weights. In their experiment, the cue association process was iterated for 200 time-steps without introducing any noise. They show that only modulatory variants of ANNs with fixed-feedforward weights and neuromodulated self-connecting recurrent weights are capable of solving this task. In our experiment, we extend a similar task to the spike domain and introduce a significant amount of sensory spike-noise. Additionally, the time dependency is more than doubled. We show that not only are neuromodulatory feedforward weights without recurrent self-connections capable of solving this task, but also that feedforward plastic weights are. We also show that the introduction of spike-noise does not decrease training convergence. On the cue association task, we show that with Oja’s rule, neuromodulatory signals drastically reduce spike-firing rates compared with the non-modulatatory variant. This reduction in activity does not apply to BCM, which has a natural mechanism for both potentiation and depression. Finally, our experiments showcase a meta-learning capability to adapt beyond what the network had encountered during its training period on a high-dimensional robotic learning task.

While our experiments showcase the performance of BCM and Oja's plasticity rules, our proposed framework can be applied to a wide variety of plasticity rules described in both the AI and neuroscience literature. Our framework may also be used to experimentally validate biological theories regarding the function of plasticity rules or neuromodulatory signals. Furthermore, the modelling of neuromodulatory signals need not be learned directly through gradient descent. Our method can be extended to explicitly model neuromodulatory signals through a pre-defined global signal. Such signals might include: an online reward signal emulating dopaminergetic neurons \cite{dopamine_hosp, dopamine_hosp}, error signals from a control system \cite{cntrl}, or a novelty signal for exploration \cite{explr}. In addition, evidence toward biological theories regarding the function of plasticity rules or neuromodulatory signals may be experimentally validated using this framework. Finally, the addition of neural processes such as homeostasis may provide further learning capabilities when interacting with differentiable synaptic plasticity. The fruitful marriage between the power of gradient descent and the adaptability of synaptic plasticity for SNNs will likely enable many interesting research opportunities for a diversity of fields. The authors see a particular enabling potential in the field of neurorobotics.

\section*{Author Contributions}
SS designed and performed the experiments as well as the analysis. SS wrote the paper with JH and JA being active contributors toward editing and revising the paper. WL also provided helpful editing of the manuscript. JH had the initial conception of the presented idea as well as having supervised the project. All authors contributed to the article and approved the submitted version.

\section*{Funding}
This work was performed at the US Naval Research Laboratory under the Base Program's Safe Lifelong Motor Learning (SLLML) work unit, WU1R36.

\section*{Supplemental Data}
It is the intent of the authors to eventually make the associated code available at:\\ \textit{\hyperlink{https://github.com/USNavalResearchLaboratory/Spikepropamine}{https://github.com/USNavalResearchLaboratory/Spikepropamine}} after it has been formally published.

\section*{Conflict of Interest Statement}

The authors declare that the research was conducted in the absence of any commercial or financial relationships that could be construed as a potential conflict of interest.

\bibliographystyle{apalike}
\bibliography{spikePropamine}

\section{Appendix}

\subsection{SLAYER Implementation}

The original code for SLAYER\footnote{https://github.com/bamsumit/slayerPytorch}, specifically the PyTorch version, was developed in CUDA to support a specific type of learning where the information regarding the temporal dataset was required to be known a-priori. In this assumed domain, a single network output was attributed to a single interaction episode, where the membrane potential was reset between these episodes. These dynamics do not support the type of flexible interactions required by RL problems, since often the interaction boundary is variable and actions must be evaluated many times before the membrane potential is reset. Toward this effort, we modified the SLAYER PyTorch library to accommodate RL applications at the low-level as well as having developed a complimentary RL framework using PPO to fit our needs. The modified SLAYER and PPO framework supports learning on the SRM model described in our experiments.

\subsection{Input Population Representation}

Specified here is the input population representation used for the high-dimensional locomotion experiment:
\begin{equation}\label{eq:POPULATIONSPROB1}
\begin{split}
& \forall m \in \{0, 1, ..., P_{dim} - 1\}, \forall \xi \in \{0, 1, ..., P_{num} - 1\}, \\
\vspace{1.3mm}
& \Omega_{\xi, m} = P_{\xi, min} (m - P_{num}) - P_{\xi, max}(P_{num} - m)
\end{split}
\vspace{0.6mm}
\end{equation}
\begin{equation}\label{eq:POPULATIONREP}
Pr[s^{(0)}_{\xi, m} = 1] = max(\vartheta_{min}, min(exp(-15 (\Omega_{\xi, m} - x_{\xi})^2), 1)).
\end{equation}
Here, the initial number of input sub-populations are defined as $P_{num} \in \mathbb{N}$ and indexed by $\xi \in \mathbb{N}$, which correspond to the number of floating point values in the pre-converted input vector. Additionally, an initial neuron population size, $P_{dim} \in \mathbb{N}$, is specified to represent each floating point input and indexed by $m \in \mathbb{N}$. Using this population, the minimum and maximum state values, $P_{\xi, max} \in \mathbb{R}$ and $P_{\xi, min} \in \mathbb{R}$, are represented by the first and last neuron in the population, and each neuron in-between is an intermediate value, $\Omega_{\xi, m} \in \mathbb{R}$, linearly distributed based on the population size (\ref{eq:POPULATIONSPROB1}). Once the place cells are appropriately represented and an incoming stimulus, $x_{\xi}$, is present, the probability of spiking for each neuron is assigned using an exponentially decaying probability distribution (\ref{eq:POPULATIONREP}). The exponential reaches its maximum value around the place neuron, $s^{(0)}_{\xi, m}$, that most closely resembles the incoming stimuli $x_{\xi}$, with each subsequent neuron in the population having a likelihood representative of the distance from this initial cell $(\Omega_{\xi, m} - x_{\xi})^2$. Additionally, a pre-defined spike probability, $\vartheta_{min} \in \mathbb{R}$, is assigned globally to each neuron independent from the distance, which was experimentally shown to improve performance on this task. We note here that the spike-activity produced by (\ref{eq:POPULATIONSPROB1}-\ref{eq:POPULATIONREP}), $s^{(0)}_{\xi, m}$, directly corresponds to the spike input defined in Section 2.

\subsection{Action Population Representation}

\begin{equation}\label{eq:POPULATIONREP22}
A_{p} = \frac{1}{T}\sum_{t=0}^{T}(\sum_{n=0}^{N} w_{n} S_{n}(t))
\end{equation}
where $A_{p} \in \mathbb{R}$ denotes the action produced over a sub-population $p \in \mathbb{N}$. The ordered-tuple of sub-population actions $A = (A_{0}, A_{1}, ..., A_{d})$ produces the final action for each actuated joint $p$, where $d \in \mathbb{N}$ is equal to the number of actuated joints. The variable $T \in \mathbb{N}$ represents the discrete time interval duration from which the action is averaged over. Additionally, $N \in \mathbb{N}$ represents the total number of neurons in the action sub-population. $S_{n}(t) \in \{0, 1\}$ is the binary spike output of neuron $n$ at time $t$, and $w_{n} \in \mathbb{R}$ weights the spike. In this experiment, $w_{n} = 1$ for half of the population $n < \frac{N}{2}$, and otherwise $w_{n} = -1$. This produces a natural mapping over the interval $[-1, 1]$ for each sub-population $A_{p}$, where simple shifting and scaling enables representation over arbitrary intervals.

\subsection{Neuron Model Hyperparameters}
Provided is a list of the neuron model hyperparameters used for the experiments in this paper. The SRMALPHA neuron type is originally described in the SLAYER code repository. We note that in practice, the behavior of the system acts independent of the defined metric units.
\begin{center}
\begin{tabular}{ |p{7.5cm}|p{2.5cm}|  }
\hline
 \multicolumn{2}{|c|}{Hyperparameter Table} \\
 \hline
 Neuron Type & SRMALPHA \\
 \hline
 Threshold & 10 (mV)\\
 \hline
 Neuron time constant & 10 (ms) \\
 \hline
 Network integration time & 1 (ms)\\
 \hline
 Refractory time constant & 2 (ms) \\
 \hline
 Neuron relative refractory response scaling & 2 \\
 \hline
 Spike function derivative time constant & 1 \\
 \hline
 Spike function derivative scale factor & 1 \\
\hline

\end{tabular}
\end{center}

\subsection{Cue-Association Training Details}

\begin{center}
\begin{tabular}{ |p{5.5cm}|p{3.8cm}|  }

 \hline
 \multicolumn{2}{|c|}{Hyperparameter Table} \\
 \hline
 Cue Labels & 2 \\
 \hline
 Total Presented Cues & 7 \\
 \hline
 Cue Presentation Time & 25 (ms)\\
 \hline
 Noise Population Neurons & 10 \\
 \hline
 Cue Population Neurons & 10 $\times$ 3 \\
 \hline
 Horizon & 500 (Steps) \\
 \hline
 Discount ($\lambda$) & 0.99 \\
 \hline
 Adam Timestep & 5 $\times 10^{-4}$ $\times$ $\alpha$ \\
 \hline
 Cue Spike Event Prob & 0.75 \\
 \hline
 Cue Spike Rest Prob & 0.05 \\
 \hline
 Noise Spike Rest Prob & 0.2 \\
 \hline
 Cue Spike Event Prob (Noisy) & 0.65 \\
 \hline
 Cue Spike Rest Prob (Noisy) & 0.25 \\
 \hline
 Noise Spike Rest Prob (Noisy) & 0.4 \\
 \hline
 Rest Period & $r \sim \{45, 75, 105\}$(ms) \\
 \hline
\end{tabular}
\end{center}

\subsection{Cue-Association: Neuronal Activity}

Here we provide additional insights for the neuronal activity during the cue-association task. The highest performing network from Experiment 1, NDP-BCM, is used for analysis.

Four cue-association cases are considered: a typical random sensory input sequence, all left cues, all right cues, and one with no sensory cues except for spike-noise. Interestingly, each of the output modulatory signal graphs exhibit a loose symmetry (Figures 6-9). The hidden modulatory signals follow a similar dynamic pattern. In the first three sequences (Figures 6-8), the weight values tend to both potentiate and depress during the cue presentation, with a stronger emphasis on negatively valued weights. The weights then seem to depress significantly during the period in which cues are absent only to potentiate again in the presence of the decision cue. However, in the fourth sequence (Figure 9), when no signals are present, the weight values do not meaningfully potentiate, and also do not depress beyond an initial range of values. The hidden modulatory activity in this case does not follow a pattern resembling the first three sequences, and seems to develop without much pattern at all. While the output modulatory signals do still exhibit a symmetry, it is the only scenario in which the signals strictly diverge from each other. It is evident that the plasticity and neuromodulatory signals have a strong effect on self-organization during the cue period, with a large population of weights undergoing drastic changes from experience-dependent activity.

\begin{figure}
\centering
\includegraphics[width=16cm]{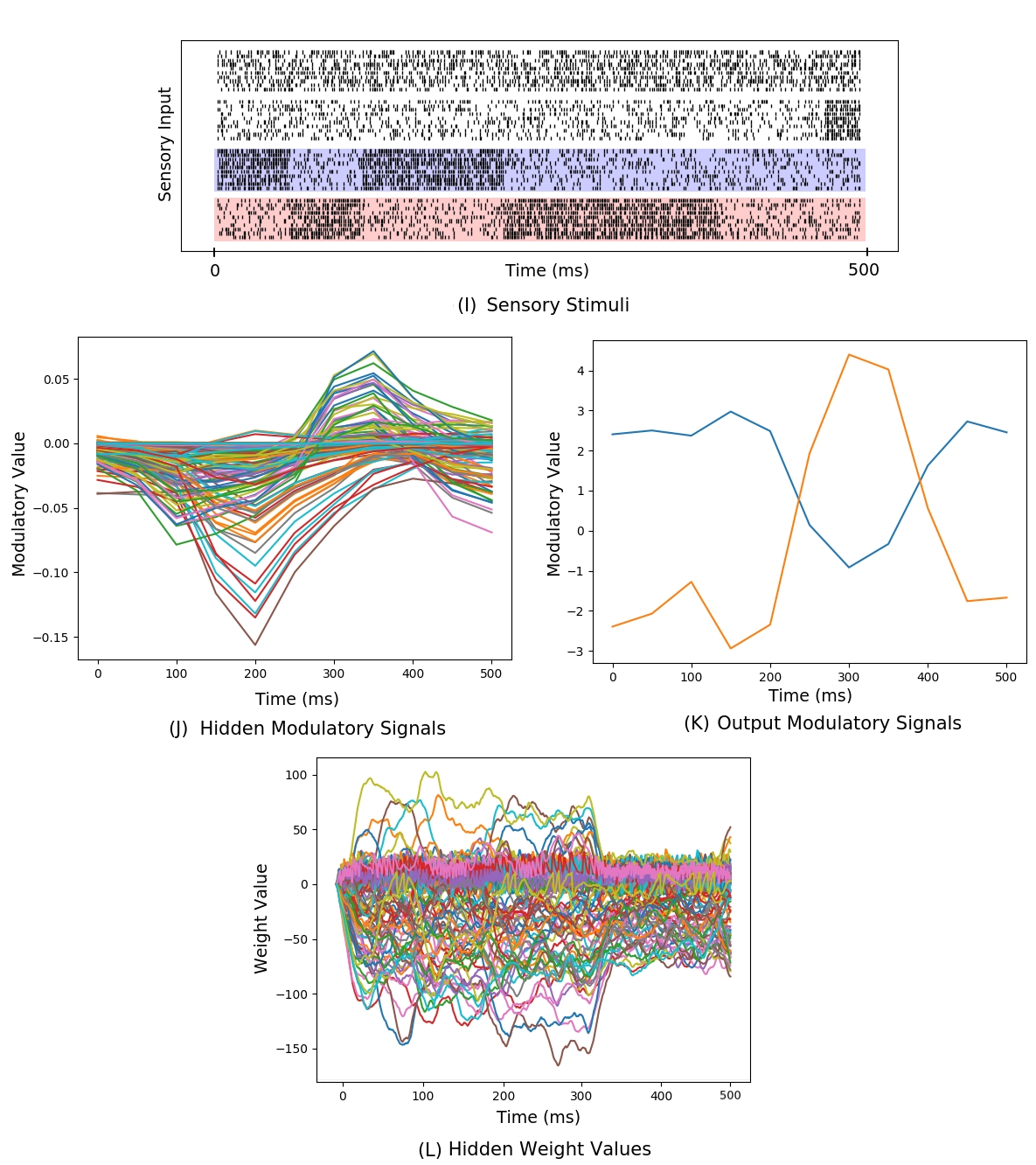}
\caption{Typical Cue Sequence}
\end{figure}

\begin{figure}
\centering
\includegraphics[width=16cm]{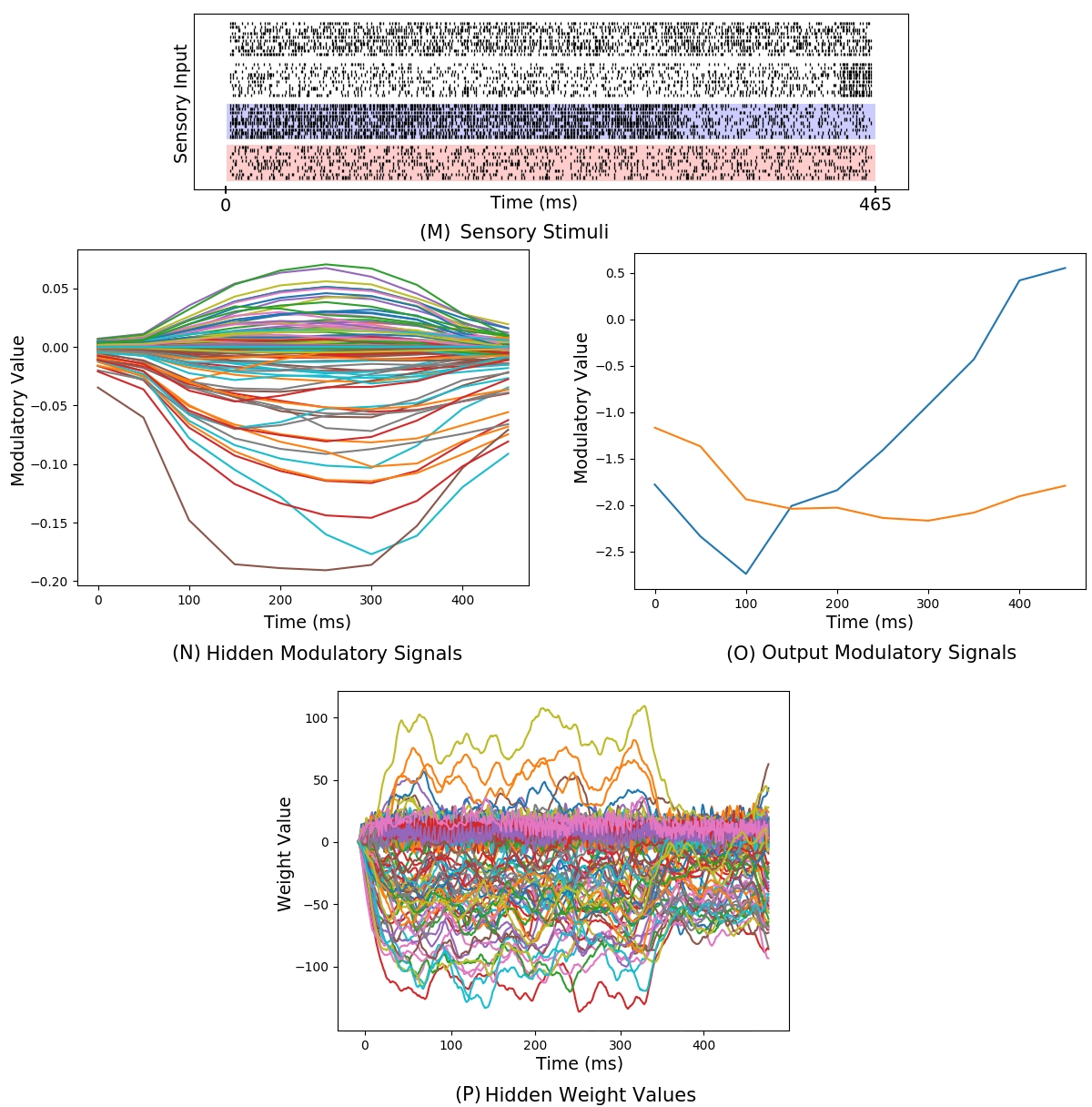}
\caption{Only Left Cues (Blue)}
\end{figure}

\begin{figure}
\centering
\includegraphics[width=16cm]{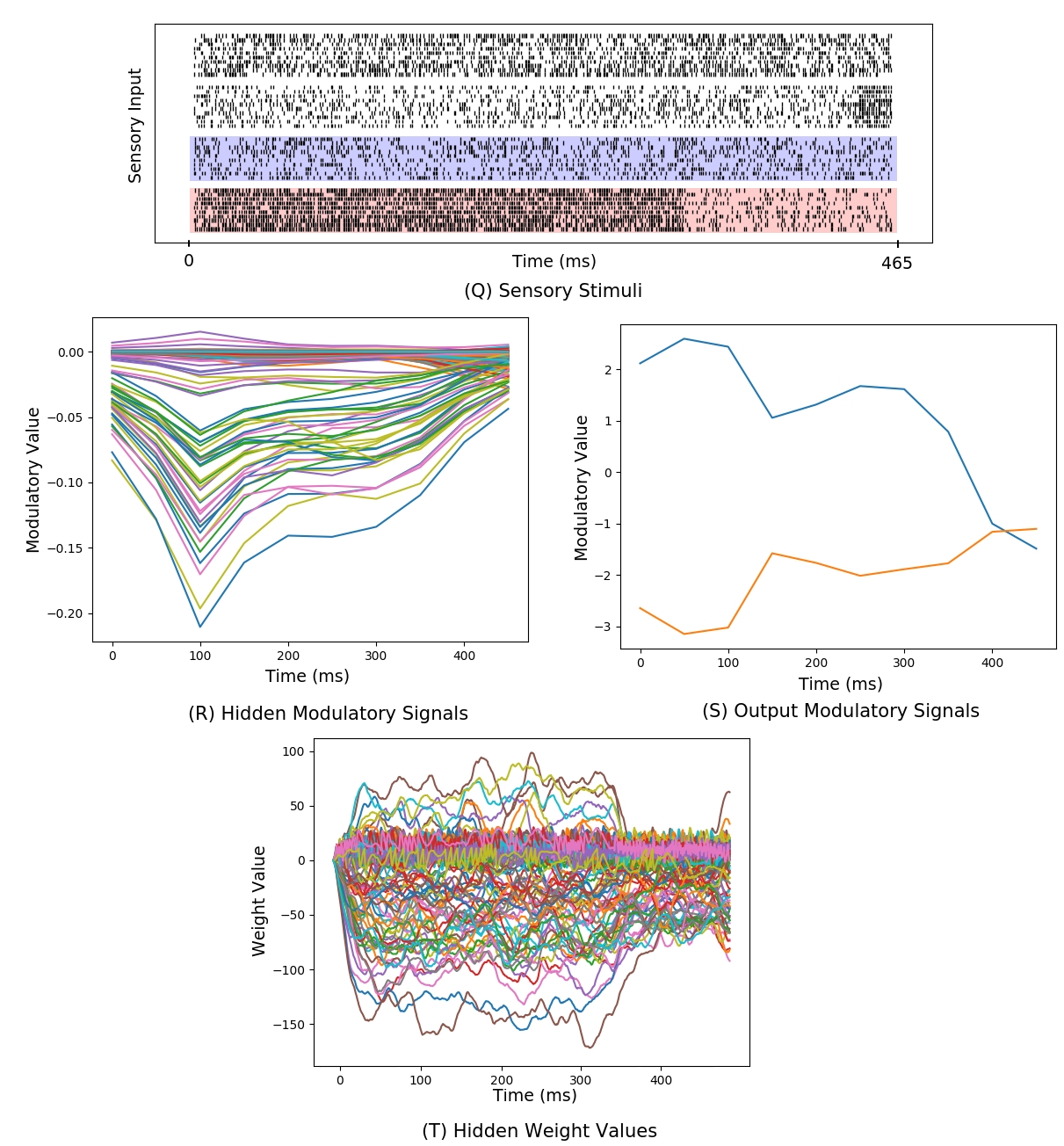}
\caption{Only Right Cues (Red)}
\end{figure}


\begin{figure}
\centering
\includegraphics[width=16cm]{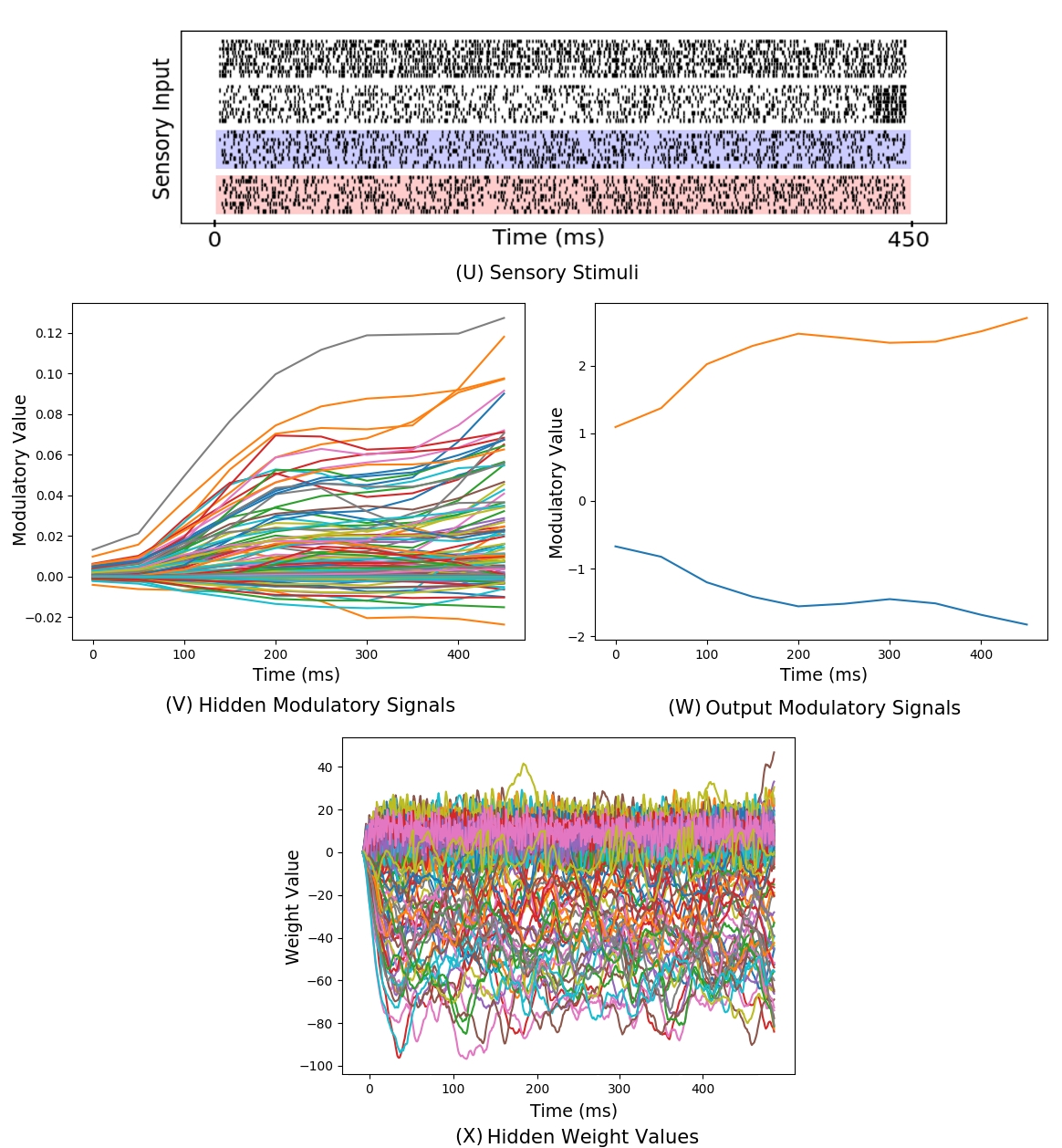}
\caption{No Sensory Input}
\end{figure}

\clearpage

\subsection{Half-Cheetah Training Details}
To compute the advantage for the Proximal Policy Optimization gradient update, Generalized Advantage Estimation (GAE) is used \cite{schulman2015highdimensional}.

%
%

\begin{center}
\begin{tabular}{ |p{5cm}|p{2.5cm}|  }

 \hline
 \hline
 \multicolumn{2}{|c|}{Hyperparameter Table} \\
 \hline
 Horizon & 3000 (Steps) \\
 \hline
 PPO Epochs & 10 \\
 \hline
 Adam Timestep & 5 $\times 10^{-4}$ $\times$ $\alpha$ \\
 \hline
 Discount ($\gamma$) & 0.99 \\
 \hline
 GAE lambda ($\lambda$) & 0.97 \\
 \hline
 PPO Updates & 1500 \\
 \hline
 Random Spike Prob $(\vartheta_{min}$) & 0.05 \\
 \hline
 Action Integration Interval (T) & 50 (ms)\\
 \hline
 \hline
\end{tabular}
\end{center}

\subsection{High-dimensional Robotic Locomotion: Neuronal Activity}

Here we provide additional insights into the internal neuronal activity for the robotic locomotion task. On this task, results are shown using with the highest-performing network from Experiment 1, NDP-Oja's.

(Figure 10 (A-D)) shows the modulatory behavior in the first hidden layer using the same network for two scenarios: when the robot is flipped on it's back (Figure 10 (A)), and when the robot is successfully performing locomotion (Figure 10 (B)). In both of these cases the action output layer is amplified by $\pm \mathcal{N}(0, 30)\%$ action noise at each timestep. In the case of successful locomotion (Figure 10 (B)), it is observed that each modulatory signal oscillates within a set region determined within 50 timesteps of the simulation. The majority of signals cluster around 0, however some signals are distributed within the range of $\pm 4$. When deprived of sensory stimuli in the flipped scenario (Figure 10 (A)) the signals still seem to display a similar distribution, however they do not exhibit nearly any oscillations. Additionally, these signals are notably larger than the hidden layer signals in the cue-association task, and do not display the same characteristic movement. Perhaps this noisy distribution plays a critical role in the adaptive behavior observed in Experiment 2.

\begin{figure}
\centering
\includegraphics[width=16cm]{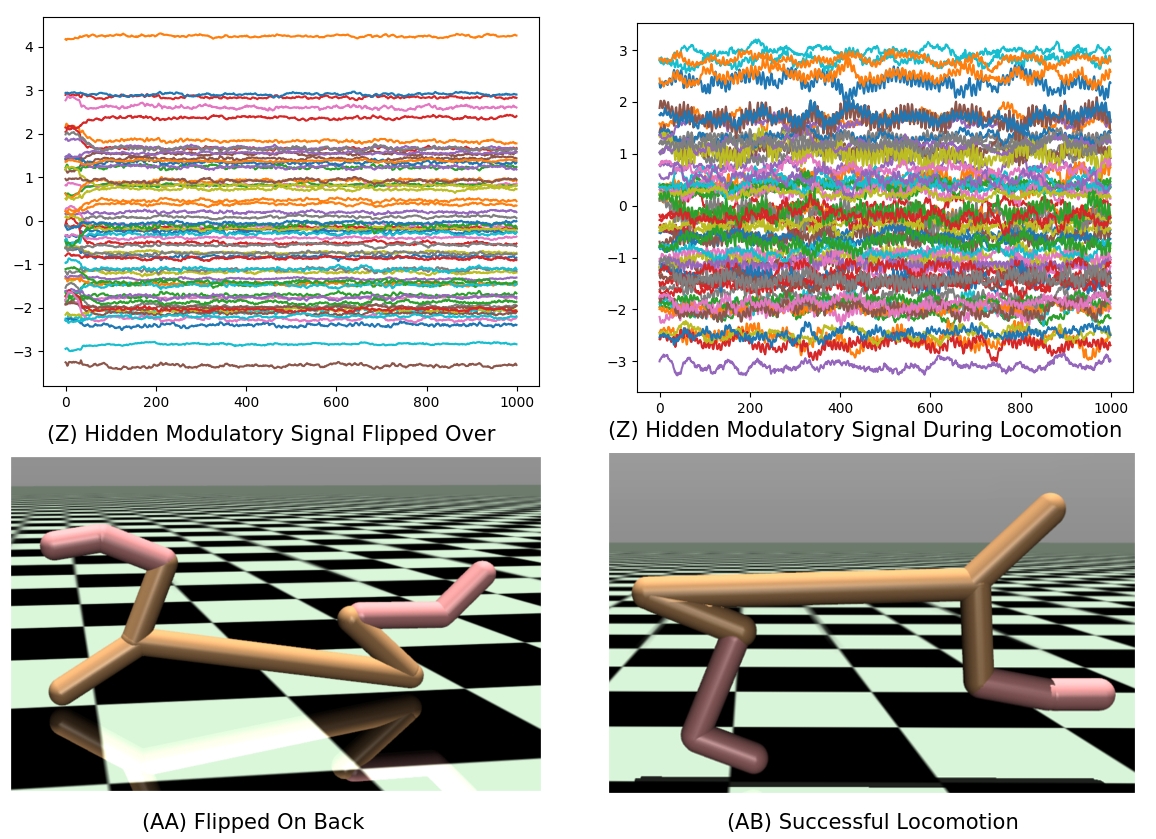}
\caption{Neuromodulatory activity in hidden layer with $\pm \mathcal{N}(0, 30)\%$ action noise when (AA) robot is flipped on back and (AB) successfully solves locomotion task. While the neuromodulatory signals across neurons seem to remain within a consistent activity region, the oscillatory behavior seems to play a role in sensory processing since, when deprived of sensory stimuli (flipped on it's back), the signals drastically reduce any change in activity.}
\end{figure}

\end{document}